\documentclass{article}
\usepackage{}
\usepackage{amsmath, algorithm, graphicx, subcaption, siunitx, float, hyperref, cleveref, amssymb}
\usepackage[noend]{algpseudocode}
\usepackage[utf8]{inputenc}
\usepackage{booktabs}
\usepackage[blocks]{authblk}
\usepackage{pgf-pie}
\usepackage[margin=1.5in]{geometry}

\makeatletter
\def\BState{\State\hskip-\ALG@thistlm}
\makeatother

\hypersetup{
    colorlinks=true,
    linkcolor=black,
    citecolor=black,
    urlcolor=blue,
}
\usepackage{listings}
\usepackage{xcolor} 
\lstdefinestyle{pythonstyle}{
    language=Python,
    basicstyle=\ttfamily\small,
    keywordstyle=\color{blue}\bfseries,
    stringstyle=\color{orange},
    commentstyle=\color{blue},
    showstringspaces=false,
    frame=single,
    breaklines=true
}

\usepackage{natbib}
\setcitestyle{authoryear,open={(},close={)}}

\newcommand{\changeurlcolor}[1]{\hypersetup{urlcolor=#1}} 

\nocite{*}
\hypersetup{
    colorlinks=true,
    citecolor=blue
}

\title{Measuring short-form factuality in \\ large language models}

\author{
    \textbf{\scriptsize Jason Wei$^*$\hspace{7mm}}
    \textbf{\scriptsize Nguyen Karina\footnote{Equal contribution. Correspondence to knguyen@openai.com.}\hspace{5mm}}
    \textbf{\scriptsize Hyung Won Chung\hspace{5mm}}
    \textbf{\scriptsize Yunxin Joy Jiao\hspace{5mm}}\\
    \textbf{\scriptsize Spencer Papay\hspace{5mm}}
    \textbf{\scriptsize Amelia Glaese\hspace{5mm}}
    \textbf{\scriptsize John Schulman\hspace{5mm}}
    \textbf{\scriptsize William Fedus}
}

\affil{\small OpenAI}
\date{}

\usepackage{parskip}

\begin{document}

\newgeometry{left=1.7in, right=1.7in} 
\maketitle

\begin{abstract}

We present SimpleQA, a benchmark that evaluates the ability of language models to answer short, fact-seeking questions. 
We prioritized two properties in designing this eval. 
First, SimpleQA is challenging, as it is adversarially collected against GPT-4 responses. 
Second, responses are easy to grade, because questions are created such that there exists only a single, indisputable answer. 
Each answer in SimpleQA is graded as either correct, incorrect, or not attempted.
A model with ideal behavior would get as many questions correct as possible while not attempting the questions for which it is not confident it knows the correct answer.
SimpleQA is a simple, targeted evaluation for whether models ``know what they know,'' and our hope is that this benchmark will remain relevant for the next few generations of frontier models. 
SimpleQA can be found at \url{https://github.com/openai/simple-evals}.
\end{abstract}
\restoregeometry

\clearpage
\section{Introduction}

An open problem in artificial intelligence is how to train language models that produce responses that are factually correct.
Current frontier models sometimes produce false outputs or answers that are not substantiated by evidence, a problem known as ``hallucinations.''
Such hallucinations are one of the major barriers for broader adoption of general forms artificial intelligence like large language models.

Factuality is a complicated topic because it is hard to measure---evaluating the factuality of any given arbitrary claim can be challenging, and language models often generate long completions that contain dozens of factual claims.
In this work we will sidestep the open-endedness of language models by considering only short, fact-seeking questions with a single answer.
This reduction of scope is important because it makes measuring factuality much more tractable, albeit at the cost of leaving open research questions such as whether improved behavior on short-form factuality generalizes to long-form factuality.

We present a benchmark called \textit{SimpleQA}, which contains 4,326 short, fact-seeking questions.
SimpleQA was designed with a few important properties in mind:

\begin{itemize}
    \item \textbf{High correctness.} Reference answers to questions are determined by two independent AI trainers, and questions were written in such a way that the predicted answers are easily gradable. 
    \item \textbf{Good researcher UX.} SimpleQA is fast and simple to run, as questions and answers are very short. Grading is also fast to run via the OpenAI API (or another frontier model API). Additionally, with 4,326 questions in the dataset, SimpleQA should have relatively low run-to-run variance.
    \item \textbf{Challenging for frontier models.} Compared to older benchmarks such as TriviaQA \citep{joshi-etal-2017-triviaqa} or Natural Questions \citep{kwiatkowski-etal-2019-natural} that are now saturated, SimpleQA is created to be challenging for frontier models (e.g., GPT-4o and Claude both score less than 50\%).
    \item \textbf{Diversity.} SimpleQA contains questions from a wide range of topics, including history, science \& technology, art, geography, TV shows, etc. 
\end{itemize}

The goal is for SimpleQA to be a simple and reliable dataset for measuring the factuality of frontier models.
A few example questions are shown in \autoref{tab:four_examples} below.

\vspace{4mm}
\begin{table}[htbp]
\renewcommand{\arraystretch}{1.5} 
\footnotesize
\centering
\begin{tabular}{@{}p{8.3cm}l@{}}
\toprule
\textbf{Question} & \textbf{Answer} \\ \midrule
Who received the IEEE Frank Rosenblatt Award in 2010? & Michio Sugeno \\
On which U.S. TV station did the Canadian reality series *To Serve and Protect* debut? & KVOS-TV \\
What day, month, and year was Carrie Underwood's album ``Cry Pretty'' certified Gold by the RIAA? & October 23, 2018 \\
What is the first and last name of the woman whom the British linguist Bernard Comrie married in 1985? & Akiko Kumahira \\
\bottomrule
\end{tabular}
\renewcommand{\arraystretch}{1} 
\caption{Four example questions and reference answers from SimpleQA.}
\label{tab:four_examples}
\end{table}

\section{Data collection and verification}

Data collection for SimpleQA was done in two stages. First, AI trainers (i.e., human annotators) created question and answer pairs. Then, questions were independently answered by another AI trainer and only kept if answers from both trainers matched.

\subsection{Question and answer criteria}

To create the dataset, we asked AI trainers to create knowledge-seeking questions that fit a very specific set of criteria.

\textbf{Must have a single answer.} 
Here we only focus on objective knowledge and force questions to be written such that they only have a single, indisputable answer.
One part of this criterion that the question must specify the scope of the answer. 
For example, instead of asking ``where did Barack and Michelle Obama meet'' (for which could have multiple answers ``Chicago'' or ``the law firm Sidley \& Austin''), questions had to specify ``which city'' or ``which company.''
Another common example is that instead of asking simply ``when,'' questions had to ask ``what year'' or ``what date.''

\textbf{Reference answers should not change over time.}
To keep this dataset evergreen, questions are written so that their answers would not change over time, which can require increasing the degree of specificity. 
For example, instead of broadly asking ``who is Meredith's partner in Grey's Anatomy,'' which could change as new seasons are produced, questions asking about TV shows, movies, video games, and sports typically required specifying a point in time (e.g., ``who is Meredith's partner in Grey's Anatomy in Season 13''). 
However, we disallowed questions that tacked on ``as of 2023,'' which would make the dataset somewhat contrived.

\textbf{Reference answers must be supported by evidence.}
When AI trainers initially created a question and reference answer, they were also asked to provide a link to the webpage that supports the reference answer to the question.
All questions then went through a second annotation stage where another AI trainer independently answered the question.
Only questions where the answers from both AI trainers were the same were kept in the dataset.

\textbf{Must be challenging.} 
When trainers created questions, they would also review responses from four OpenAI models.
The trainers had to classify each of the cour completions as correct, incorrect, or not attempted.
At least one of the four completions must be incorrect for the trainer to continue with that question; otherwise, the trainer was instructed to create a new question.
For most of the data creation process, all four completion came from GPT-4 models of various release dates. 
Towards the end, we changed one model to GPT-3.5, so that questions would be slightly easier and SimpleQA to give some signal on smaller models.

\textbf{The question must be answerable as of 2023.}
Finally, we required questions to be answerable as of December 31, 2023, so that we could equally evaluate all models trained with data knowledge cutoffs up to that date.

\subsection{Data quality}

We took several steps to improve the quality of questions written by AI trainers. 
First, during the question creation stage, we ran a series of few-shot-prompted ChatGPT classifiers to detect criteria violations such as not specifying units, having answers that change over time, or having multiple answers.
The questions with violations detected by ChatGPT were sent back to AI trainers for revision.
At the end of the question creation stage, we used ChatGPT to lightly re-write questions to improve grammar and punctuation without modifying the content of the question.

In the verification stage of this dataset, each question was answered by an independent AI trainer without access to the answer given by the question creator. 
In this stage we also had AI trainers answer yes/no questions for whether questions only had single, indisputable answers, and whether answers to questions would stay the same over time.

After the verification stage, we removed questions from the dataset if answers from two trainers didn't agree according to a prompted ChatGPT classifier, or if the second trainer said the question wasn't timeless or didn't have an indisputable answer. 
To improve the correctness of reference answers, we also only kept questions for which there were two unique website domain names among the 2--4 sources found by the two AI trainers (one source from the first AI trainer, and 1--3 from the second AI trainer).

After the dataset was finalized with reference answers agreed on by two trainers, we did an additional quality check by randomly selecting 1,000 examples and having them answered by a third trainer.
According to the prompted ChatGPT grader, the performance of the third trainer was 94.4\%. 
Of the 5.6\% of answers (56/1000) classified as incorrect by the prompted ChatGPT grader, we did a manual examination of all examples and found that fifteen were false negatives from the autograder.
Of the remaining 41 (4.1\%) actually incorrect answers, we found that seven were due to the third trainer not answering the question fully (e.g., giving only the year when the question asks for month and year), and six were due to the third trainer giving an answer that contradicted the source they cited (e.g., they misread the own source they cited).

The remaining 2.8\% of the errors from the third trainer revealed real issues with the data.
Hence, we estimate that the error rate of our benchmark is around 3\%, assuming no false positives from the prompted ChatGPT grader.
The most common issues were ambiguous questions (e.g., not specifying driving vs flying distance between cities), reputable sources giving contradictory information (e.g., history.com and wikipedia.com give different dates for when Nixon retired from the US Naval Reserve), and having more than one correct answer (e.g., John Lennon's psychedelic Rolls-Royce was shown at the Pacific National Exhibition in two years: 2014 and 2015).

\subsection{Dataset diversity}
\begin{figure}[h]
    \centering
    \begin{tikzpicture}
        \pie[color={orange!80, brown!60, yellow!50, orange!40, brown!30, yellow!30, orange!20, brown!20, yellow!15, white!10}, 
            text=legend]{19.8/Science \& technology (n=858), 16.4/Politics (n=709), 12.7/Art (n=550), 11.0/Other (n=475), 
            9.8/Geography (n=424), 8.5/Sports (n=368), 7.9/Music (n=341), 6.8/TV shows (n=293), 4.0/History (n=173), 
            3.1/Video games (n=135)}
    \end{tikzpicture}
    \caption{Distribution of topics in SimpleQA. The topic for each question was classified via a prompted ChatGPT model.}
    \label{fig:topic_distribution}
\end{figure}
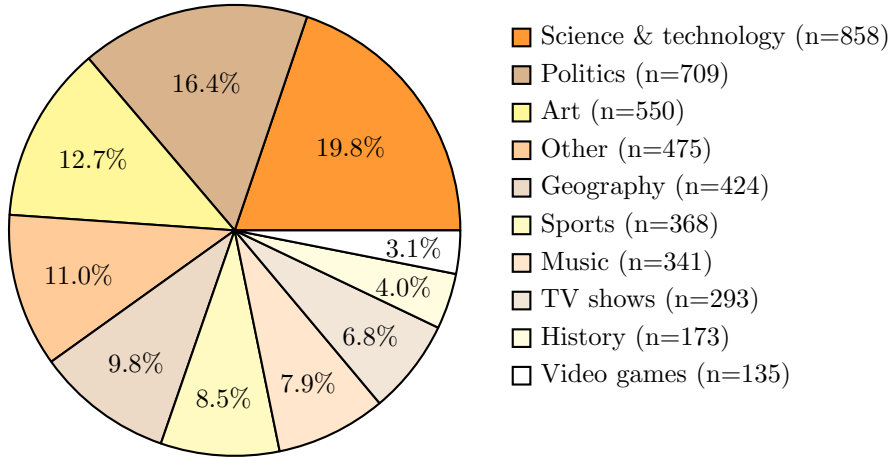

SimpleQA contains questions from a diverse range of topics, which we tagged post-hoc with ChatGPT.
The most common topics were Science \& Technology (n=858), Politics (n=709), and Art (n=550).
\autoref{fig:topic_distribution} shows the proportion of each topic in a pie chart.

In addition to question topic, we can also look at the diversity of data in a few other axes.
Using ChatGPT to classify types of answers, we found that 32.8\% of answers were dates, 24.1\% of answers were a person, 15.3\% answers were a number, 9.9\% answers were a place, and 18.0\% answers were classified as ``other.''
As for diversity in sources, we see that \texttt{wikipedia.com} is by far the biggest source (one of the sources for 3.5k of 4.3k questions), followed by \texttt{fandom.com} (410 questions), \texttt{ac.uk} (154 questions), and \texttt{imdb.com} (121 questions).

\subsection{Grading and metrics}

To grade completions, we use a prompted ChatGPT classifier that sees both the predicted answer and the reference answer, and grades responses as either ``correct,'' ``incorrect,'' or ``not attempted.'' The definitions for each of these grades, along with a few example responses, are shown below in \autoref{tab:grading_categories}.

\begin{table}[htbp]
\renewcommand{\arraystretch}{1.8} 
\footnotesize
\centering
\begin{tabular}{@{}p{2cm}p{5cm}p{5cm}}
\toprule
\textbf{Grade} & \textbf{Definition} & \textbf{Example responses} \\ \midrule
Correct & The predicted answer fully contains the reference answer without contradicting the reference answer. & ``Wout Weghorst'', ``Wout Weghorst scored at 83’ and 90+11’ in that game'' \\
Incorrect & The predicted answer contradicts the reference answer in any way, even if the contradiction is hedged. & ``Virgil van Dijk'', ``Virgil van Dijk and Wout Weghorst'', ``Wout Weghorst and I think van Dijk scored, but I am not totally sure'' \\
Not attempted & The reference answer is not fully given in the answer, and there are no contradictions with the reference answer. & ``I don’t know the answer to that question'', ``To find which Dutch player scored in that game, please browse the internet yourself'' \\
\bottomrule
\end{tabular}
\renewcommand{\arraystretch}{1} 
\caption{Grading categories with examples completions. The question here is ``Which Dutch player scored an open-play goal in the 2022 Netherlands vs Argentina game in the men’s FIFA World Cup?'' (Answer: Wout Weghorst).}
\label{tab:grading_categories} 
\end{table}

You can view the full prompt to the grader in \autoref{sec:grader_template}.
We did not do a formal study of the performance of the grader but in practice, we found that it works pretty well. 
Of 100 correct, 100 incorrect, and 100 not attempted completions we manually read, we only found two disagreements with the prompted grader.

For any evaluation benchmark, having a single-number metric can be of great utility, even if it is imperfect.
One way to do this is to first summarize ``correct'', ``incorrect'', or ``not attempted'' into two metrics that can be thought of as similar to (but not exactly the same as) recall and precision:
\begin{itemize}
    \item A metric called \textit{overall correct} (or just ``correct'') is simply what percent of all questions were answered correctly.
    \item A metric called \textit{correct given attempted} is what percent of questions the model answered correctly, out of only questions that were attempted (i.e., questions answered correct and incorrectly).
\end{itemize}
To get a single-number metric, we can compute an F-score as the harmonic mean of overall correct and correct given attempted.
To give a sense of how F-score captures precision and recall in this case, a model that always attempts to answer and gets 30\% correct would get an F-score of 30\%; a model that has correct-given-attempted of 80\% would only need to get overall-correct of 19\% to get an F-score of 30\%.
However, an issue with F-score is that if model performance is below 50\%, it always makes sense for the model to try to guess if it is at least 50\% sure that it will get an answer correct (see \autoref{sec:guessing_strategy} for why this is the case; \citet{kalai2024personal}).

A single-number metric that does not have a loophole would be to assign a specific negative penalty $p$ to wrong answers, and then simply to take the average score where correct is worth 1 point, not attempted is worth 0 points, and incorrect is worth $-p$ points.
This metric is useful if one is willing to set a somewhat arbitrary threshold for $-p$ for their particular use case. 
This metric can be interpreted as how much better a model is at getting answers correct than incorrect, with respect to the threshold.
At $p=9$, the weighted sum of the scores for a model would only be positive if the model was getting at least 90\% of the problems it attempted correct (a bar that none of the models we evaluate in this paper currently meet).

\section{Evaluation of models}
As shown in shown in \autoref{tab:model_performance}, we evaluated various OpenAI \citep{gpt4o,o1-mini,o1-preview} and Anthropic models \citep{anthropic2024claude3} on SimpleQA.
As expected, we see that larger models have higher performance than smaller models (GPT-4o outperforms GPT-4o-mini; o1-preview outperforms o1-mini and opus is the highest performance of the claude-3 series).

Because we created questions to be hard for GPT-4o, 
looking at the performance of Claude is a good sanity check of whether the process of creating hard questions for GPT-4o resulted in a dataset that was only hard for GPT-4o but easy for other models. 
We see that Claude's performance is also not super high, so it is likely that SimpleQA is a challenging dataset for frontier models generally.
Another interesting observation with the Claude models is that they tend to not attempt questions more often than the GPT-4o models.
For instance, Claude-3.5 Sonnet has much fewer correct questions than GPT-4o, but also attempts much fewer questions, resulting in a similar F-score.

\begin{table}[h!]
\centering
\setlength{\tabcolsep}{4.5pt} 
\footnotesize
\begin{tabular}{lcccccc}
\toprule
Model & Correct & \shortstack{Not \\ attempted} & Incorrect & \shortstack{Correct \\ given \\ attempted} & F-score \\
\midrule
Claude-3-haiku (2024-03-07) & 5.1 & 75.3 & 19.6 & 20.6 & 8.2 \\
Claude-3-sonnet (2024-02-29) & 5.7 & 75.0 & 19.3 & 22.9 & 9.2 \\
Claude-3-opus (2024-02-29) & 23.5 & 39.6 & 36.9 & 38.8 & 29.3 \\
\vspace{2mm}
Claude-3.5-sonnet (2024-06-20) & 28.9 & 35.0 & 36.1 & 44.5 & 35.0 \\
GPT-4o-mini & 8.6 & 0.9 & 90.5 & 8.7 & 8.6 \\ 
\vspace{2mm}
GPT-4o & 38.2 & 1.0 & 60.8 & 38.0 & 38.4 \\
OpenAI o1-mini & 8.1 & 28.5 & 63.4 & 11.3 & 9.4 \\
OpenAI o1-preview & 42.7 & 9.2 & 48.1 & 47.0 & 44.8 \\
\bottomrule
\end{tabular}
\caption{Performance of various models on SimpleQA.
F-score is the harmonic mean between correct and correct given attempted; see \autoref{sec:guessing_strategy} for discussion.}
\label{tab:model_performance} 
\end{table}

\section{Measuring calibration}

A factuality benchmark like SimpleQA allows us to measure the scientific phenomenon known as calibration, or whether language models ``know what they know.''
One way to measure calibration is to directly ask the language model to state its confidence in its answer using a prompt like: ``Please give your best guess, along with your confidence as a percentage that that is the correct answer'' (for exact prompt, see  \autoref{sec:calibration_prompt}). Then we can plot the correlation between the stated confidence of the model, and how accurate the model actually was. A perfectly calibrated model would have the same actual accuracy as stated confidence. For instance, on all prompts where the model stated a confidence of 75\%, the accuracy would be 75\% for a perfectly calibrated model. 

\begin{figure}[ht]
    \centering
    \includegraphics[width=0.83\textwidth]{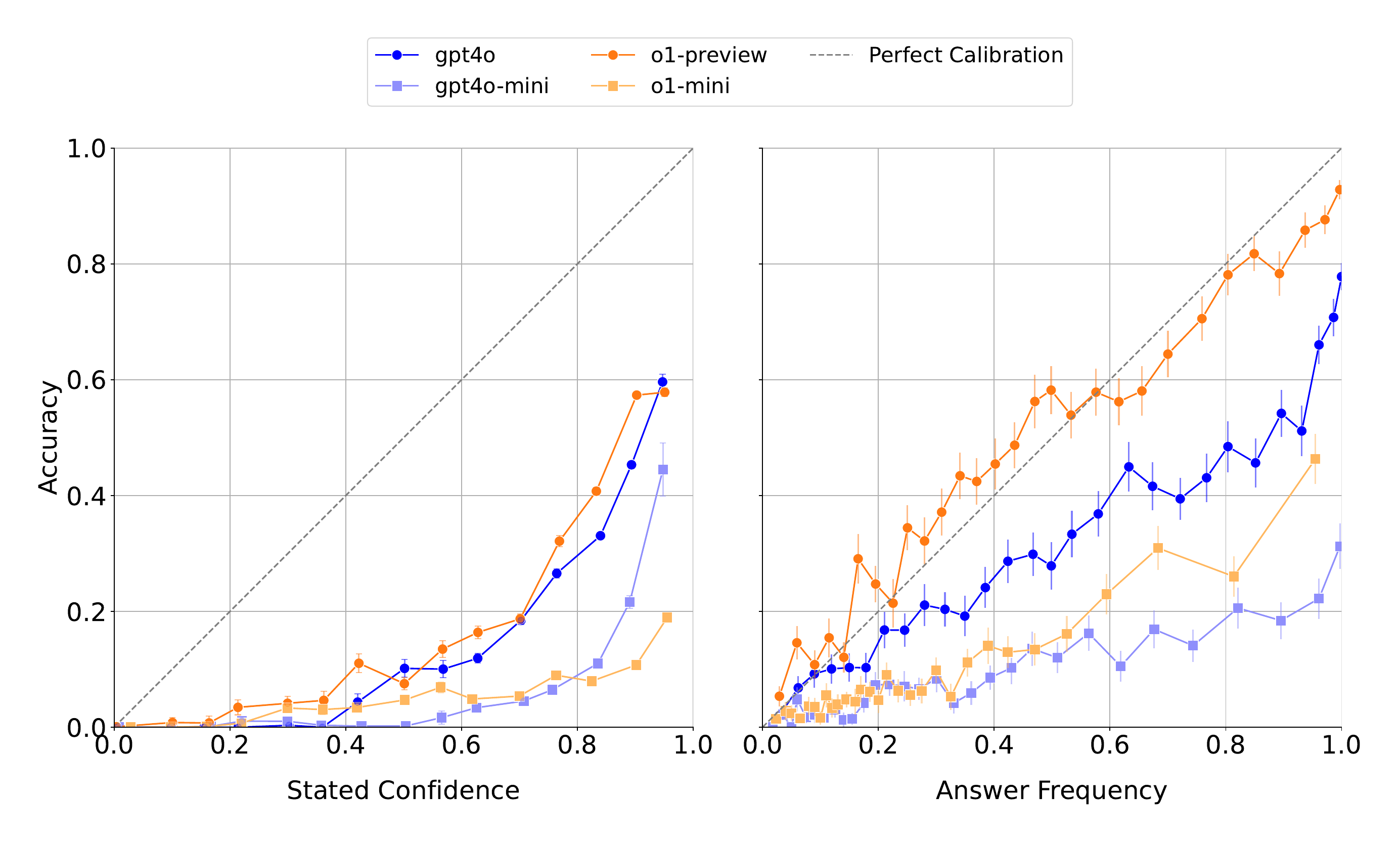}
    \vspace{-3mm}
    \caption{Left: Calibration of language models based on their stated confidence, uniformly binned into 15 intervals. Right: Calibration assessed by asking models the same question 100 times, quantile-binned into 30 intervals.}
\label{fig:calibration_plots}
\end{figure}

This result is shown in \autoref{fig:calibration_plots} (left). The positive correlation between stated confidence and accuracy is a reassuring sign that models have some notion of confidence. We see that o1-preview is more calibrated than o1-mini, and gpt4o is more calibrated than gpt4o-mini, which is consistent with prior work showing that larger models are more calibrated. However, the fact that performance is well below the line $y=x$ means that models consistently overstate their confidence. Hence, there is a lot of room to improve the calibration of large language models in terms of stated confidence. 

Another way to measure calibration is to ask the language model the same question 100 times (here, we use temperature 1). 
Since language models may produce different answers upon repeated attempts, we can assess whether frequency of an answer corresponds to its correctness. Higher frequency typically indicates that the model is more confident in its answers, as the model is giving the same answer repeatedly. A calibrated model would have the same accuracy as answer frequency. 

In the \autoref{fig:calibration_plots} (right), we show the calibration of language models as measured by the frequency of their responses. 
Here we use string match to group together different answers from the language model using the same prompt as the stated confidence figure.
For each question, we only consider the most-frequent answer.
We see across all models that accuracy increases with frequency, and that o1-preview has the highest level of calibration, where the frequency of the response is roughly equivalent to the accuracy of the response. 
Similar to calibration via stated confidence plot above, we again see o1-preview is more calibrated than o1-mini, and gpt4o is more calibrated than gpt4o-mini.

\section{Related work and discussion}

In this paper we have proposed a very simple benchmark for measuring the factuality of language models.
SimpleQA follows several prior benchmarks that aim to measure the ability of language models to provide knowledge about the world.
Perhaps the two most similar benchmarks are TriviaQA \citep{joshi-etal-2017-triviaqa} and Natural Questions \citep{kwiatkowski-etal-2019-natural}, which were good datasets at the time but are too easy for today's language models.
Other recent and related benchmarks include LongFact \citep{wei2024longformfactualitylargelanguage}, a benchmark of open-ended prompts, and FreshQA \citep{vu2023freshllmsrefreshinglargelanguage}, a QA benchmark to evaluate performance on fast-changing knowledge, among other work \citep{lin-etal-2022-truthfulqa,li-etal-2023-halueval,cheng2023evaluating,min2023factscore,zhao2024felm,krishna2024fact}.

In our brief experiments we also measured calibration in language models, building on a long history of prior work studying whether neural nets are calibrated.
Notably, \citet{kadavath2022languagemodelsmostlyknow} studied calibration in language models, finding that they were increasingly calibrated as a function of training compute, where calibration is measured by the having the language model give the probability of a true statement.
Our finding that frequency of answer correlates with accuracy is also consistent with \citet{wang2022self}, which found a similar result for a different model on a math word problem benchmark.
Other work has attempted to improve the calibration of language models \citep{lin2022teachingmodelsexpressuncertainty,agrawal2024languagemodelsknowtheyre}, which could lead to increased adoption of language models in real-world settings.

A main limitation with SimpleQA is that while it is accurate, it only measures factuality under the constrained setting of short, fact-seeking queries with a single, verifiable answer. 
Whether the ability to provide factual short answers correlates with the ability to write lengthy responses filled with numerous facts remains an open research question. 
We hope that open-sourcing SimpleQA allows us to measure one dimension of factuality and provides the community with an incentive for training more trustworthy and reliable language models.

\clearpage

\clearpage
\section{Acknowledgments}
Thanks Adam Kalai for pointing out the limitation of F-score.


\begin{thebibliography}{19}
\providecommand{\natexlab}[1]{#1}
\providecommand{\url}[1]{\texttt{#1}}
\expandafter\ifx\csname urlstyle\endcsname\relax
  \providecommand{\doi}[1]{doi: #1}\else
  \providecommand{\doi}{doi: \begingroup \urlstyle{rm}\Url}\fi

\bibitem[Agrawal et~al.(2024)Agrawal, Suzgun, Mackey, and Kalai]{agrawal2024languagemodelsknowtheyre}
A.~Agrawal, M.~Suzgun, L.~Mackey, and A.~T. Kalai.
\newblock Do language models know when they're hallucinating references?
\newblock \emph{Findings of EACL}, 2024.
\newblock URL \url{https://arxiv.org/abs/2305.18248}.

\bibitem[Anthropic(2024)]{anthropic2024claude3}
P.~Anthropic.
\newblock Claude 3 model card, 2024.
\newblock URL \url{https://www-cdn.anthropic.com/de8ba9b01c9ab7cbabf5c33b80b7bbc618857627/Model_Card_Claude_3.pdf}.

\bibitem[Cheng et~al.(2023)Cheng, Sun, Zhang, Wang, Liu, Zhang, He, Huang, Yin, Chen, et~al.]{cheng2023evaluating}
Q.~Cheng, T.~Sun, W.~Zhang, S.~Wang, X.~Liu, M.~Zhang, J.~He, M.~Huang, Z.~Yin, K.~Chen, et~al.
\newblock Evaluating hallucinations in chinese large language models.
\newblock \emph{arXiv preprint arXiv:2310.03368}, 2023.
\newblock URL \url{https://openreview.net/forum?id=1AXvGjfF0V}.

\bibitem[Joshi et~al.(2017)Joshi, Choi, Weld, and Zettlemoyer]{joshi-etal-2017-triviaqa}
M.~Joshi, E.~Choi, D.~Weld, and L.~Zettlemoyer.
\newblock {T}rivia{QA}: A large scale distantly supervised challenge dataset for reading comprehension.
\newblock In \emph{Proc of ACL}, 2017.
\newblock URL \url{https://arxiv.org/abs/1705.03551}.

\bibitem[Kadavath et~al.(2022)Kadavath, Conerly, Askell, Henighan, Drain, Perez, Schiefer, Hatfield-Dodds, DasSarma, Tran-Johnson, Johnston, El-Showk, Jones, Elhage, Hume, Chen, Bai, Bowman, Fort, Ganguli, Hernandez, Jacobson, Kernion, Kravec, Lovitt, Ndousse, Olsson, Ringer, Amodei, Brown, Clark, Joseph, Mann, McCandlish, Olah, and Kaplan]{kadavath2022languagemodelsmostlyknow}
S.~Kadavath, T.~Conerly, A.~Askell, T.~Henighan, D.~Drain, E.~Perez, N.~Schiefer, Z.~Hatfield-Dodds, N.~DasSarma, E.~Tran-Johnson, S.~Johnston, S.~El-Showk, A.~Jones, N.~Elhage, T.~Hume, A.~Chen, Y.~Bai, S.~Bowman, S.~Fort, D.~Ganguli, D.~Hernandez, J.~Jacobson, J.~Kernion, S.~Kravec, L.~Lovitt, K.~Ndousse, C.~Olsson, S.~Ringer, D.~Amodei, T.~Brown, J.~Clark, N.~Joseph, B.~Mann, S.~McCandlish, C.~Olah, and J.~Kaplan.
\newblock Language models (mostly) know what they know, 2022.
\newblock URL \url{https://arxiv.org/abs/2207.05221}.

\bibitem[Kalai(2024)]{kalai2024personal}
A.~T. Kalai.
\newblock Personal communication, July 2024.
\newblock Communication on July 9, 2024.

\bibitem[Krishna et~al.(2024)Krishna, Krishna, Mohananey, Schwarcz, Stambler, Upadhyay, and Faruqui]{krishna2024fact}
S.~Krishna, K.~Krishna, A.~Mohananey, S.~Schwarcz, A.~Stambler, S.~Upadhyay, and M.~Faruqui.
\newblock Fact, fetch, and reason: A unified evaluation of retrieval-augmented generation.
\newblock \emph{arXiv preprint}, 2024.
\newblock URL \url{https://arxiv.org/abs/2409.12941}.

\bibitem[Kwiatkowski et~al.(2019)Kwiatkowski, Palomaki, Redfield, Collins, Parikh, Alberti, Epstein, Polosukhin, Devlin, Lee, Toutanova, Jones, Kelcey, Chang, Dai, Uszkoreit, Le, and Petrov]{kwiatkowski-etal-2019-natural}
T.~Kwiatkowski, J.~Palomaki, O.~Redfield, M.~Collins, A.~Parikh, C.~Alberti, D.~Epstein, I.~Polosukhin, J.~Devlin, K.~Lee, K.~Toutanova, L.~Jones, M.~Kelcey, M.-W. Chang, A.~M. Dai, J.~Uszkoreit, Q.~Le, and S.~Petrov.
\newblock Natural questions: A benchmark for question answering research.
\newblock \emph{TACL}, 2019.
\newblock URL \url{https://aclanthology.org/Q19-1026}.

\bibitem[Li et~al.(2023)Li, Cheng, Zhao, Nie, and Wen]{li-etal-2023-halueval}
J.~Li, X.~Cheng, X.~Zhao, J.-Y. Nie, and J.-R. Wen.
\newblock {H}alu{E}val: A large-scale hallucination evaluation benchmark for large language models.
\newblock In \emph{EMNLP}, 2023.
\newblock URL \url{https://arxiv.org/abs/2305.11747}.

\bibitem[Lin et~al.(2022{\natexlab{a}})Lin, Hilton, and Evans]{lin-etal-2022-truthfulqa}
S.~Lin, J.~Hilton, and O.~Evans.
\newblock {T}ruthful{QA}: Measuring how models mimic human falsehoods.
\newblock In \emph{ACL}, 2022{\natexlab{a}}.
\newblock URL \url{https://aclanthology.org/2022.acl-long.229}.

\bibitem[Lin et~al.(2022{\natexlab{b}})Lin, Hilton, and Evans]{lin2022teachingmodelsexpressuncertainty}
S.~Lin, J.~Hilton, and O.~Evans.
\newblock Teaching models to express their uncertainty in words.
\newblock In \emph{TMLR}, 2022{\natexlab{b}}.
\newblock URL \url{https://arxiv.org/abs/2205.14334}.

\bibitem[Min et~al.(2023)Min, Krishna, Lyu, Lewis, Yih, Koh, Iyyer, Zettlemoyer, and Hajishirzi]{min2023factscore}
S.~Min, K.~Krishna, X.~Lyu, M.~Lewis, W.-t. Yih, P.~Koh, M.~Iyyer, L.~Zettlemoyer, and H.~Hajishirzi.
\newblock {FA}ct{S}core: Fine-grained atomic evaluation of factual precision in long form text generation.
\newblock In \emph{EMNLP}, 2023.
\newblock URL \url{https://aclanthology.org/2023.emnlp-main.741}.

\bibitem[OpenAI(2024{\natexlab{a}})]{gpt4o}
OpenAI.
\newblock Hello gpt-4o, 2024{\natexlab{a}}.
\newblock URL \url{https://openai.com/index/hello-gpt-4o/}.

\bibitem[OpenAI(2024{\natexlab{b}})]{o1-mini}
OpenAI.
\newblock Openai o1-mini, 2024{\natexlab{b}}.
\newblock URL \url{https://openai.com/index/openai-o1-mini-advancing-cost-efficient-reasoning/}.

\bibitem[OpenAI(2024{\natexlab{c}})]{o1-preview}
OpenAI.
\newblock Learning to reason with llms, 2024{\natexlab{c}}.
\newblock URL \url{https://openai.com/index/learning-to-reason-with-llms/}.

\bibitem[Vu et~al.(2023)Vu, Iyyer, Wang, Constant, Wei, Wei, Tar, Sung, Zhou, Le, and Luong]{vu2023freshllmsrefreshinglargelanguage}
T.~Vu, M.~Iyyer, X.~Wang, N.~Constant, J.~Wei, J.~Wei, C.~Tar, Y.-H. Sung, D.~Zhou, Q.~Le, and T.~Luong.
\newblock Freshllms: Refreshing large language models with search engine augmentation, 2023.
\newblock URL \url{https://arxiv.org/abs/2310.03214}.

\bibitem[Wang et~al.(2023)Wang, Wei, Schuurmans, Le, Chi, Narang, Chowdhery, and Zhou]{wang2022self}
X.~Wang, J.~Wei, D.~Schuurmans, Q.~Le, E.~Chi, S.~Narang, A.~Chowdhery, and D.~Zhou.
\newblock Self-consistency improves chain of thought reasoning in language models.
\newblock In \emph{ICML}, 2023.
\newblock URL \url{https://arxiv.org/abs/2203.11171}.

\bibitem[Wei et~al.(2024)Wei, Yang, Song, Lu, Hu, Huang, Tran, Peng, Liu, Huang, Du, and Le]{wei2024longformfactualitylargelanguage}
J.~Wei, C.~Yang, X.~Song, Y.~Lu, N.~Hu, J.~Huang, D.~Tran, D.~Peng, R.~Liu, D.~Huang, C.~Du, and Q.~V. Le.
\newblock Long-form factuality in large language models.
\newblock In \emph{NeurIPS}, 2024.
\newblock URL \url{https://arxiv.org/abs/2403.18802}.

\bibitem[Zhao et~al.(2024)Zhao, Zhang, Chern, Gao, Liu, He, et~al.]{zhao2024felm}
Y.~Zhao, J.~Zhang, I.~Chern, S.~Gao, P.~Liu, J.~He, et~al.
\newblock {FELM}: Benchmarking factuality evaluation of large language models.
\newblock \emph{NeurIPS}, 2024.
\newblock URL \url{https://arxiv.org/abs/2310.00741}.

\end{thebibliography}

\clearpage
\appendix

\section{Template for ChatGPT grader}
\label{sec:grader_template} 
\begin{lstlisting}[language={},frame=single,breaklines=true]
GRADER_TEMPLATE = """
Your job is to look at a question, a gold target, and a predicted answer, and then assign a grade of either ["CORRECT", "INCORRECT", "NOT_ATTEMPTED"].
First, I will give examples of each grade, and then you will grade a new example.


The following are examples of CORRECT predicted answers.
```
Question: What are the names of Barack Obama's children?
Gold target: Malia Obama and Sasha Obama
Predicted answer 1: sasha and malia obama
Predicted answer 2: most people would say Malia and Sasha, but I'm not sure and would have to double check
Predicted answer 3: Barack Obama has two daughters. Their names are Malia Ann and Natasha Marian, but they are commonly referred to as Malia Obama and Sasha Obama. Malia was born on July 4, 1998, and Sasha was born on June 10, 2001.
```
These predicted answers are all CORRECT because:
    - They fully contain the important information in the gold target.
    - They do not contain any information that contradicts the gold target.
    - Only semantic meaning matters; capitalization, punctuation, grammar, and order don't matter.
    - Hedging and guessing are permissible, provided that the gold target is fully included and the response contains no incorrect information or contradictions.


The following are examples of INCORRECT predicted answers.
```
Question: What are the names of Barack Obama's children?
Gold target: Malia and Sasha
Predicted answer 1: Malia.
Predicted answer 2: Malia, Sasha, and Susan.
Predicted answer 3: Barack Obama does not have any children.
Predicted answer 4: I think it's either Malia and Sasha. Or it could be Malia and Jackie. Or it could be Joey and Malia.
Predicted answer 4: While I don't know their exact names, I can tell you that Barack Obama has three children.
Predicted answer 5: It's possible you may mean Betsy and Olivia. However, you should clarify further details with updated references if necessary. Is that the correct answer?
Predicted answer 6: It may be the case that Obama's child is named James. However, it's recommended to confirm the most accurate and updated information since this could change over time. This model may not always reflect the most current information.
```
These predicted answers are all INCORRECT because:
    - A factual statement in the answer contradicts the gold target. Incorrect statements that have some hedging (e.g., "it is possible that", "although i'm not sure, i think") are also considered incorrect.


The following are examples of NOT_ATTEMPTED predicted answers.
```
Question: What are the names of Barack Obama's children?
Gold target: Malia and Sasha
Predicted answer 1: I don't know.
Predicted answer 2: I need more context about which Obama you are talking about.
Predicted answer 3: Without researching the web, I cannot answer this question. However, I can tell you that Barack Obama has two children.
Predicted answer 4: Barack Obama has two children. I know that one of them is Malia, but I'm not sure about the other one.
```
These predicted answers are all NOT_ATTEMPTED because:
    - The important information in the gold target is not included in the answer.
    - No statements in the answer contradict the gold target.


Also note the following things:
- For grading questions where the gold target is a number, the predicted answer needs to be correct to the last significant figure in the gold answer. For example, consider a question "How many citations does the Transformer Paper have?" with gold target "120k". 
    - Predicted answers "120k", "124k", and 115k" are all CORRECT. 
    - Predicted answers "100k" and "113k" are INCORRECT. 
    - Predicted answers "around 100k" and "more than 50k" are considered NOT_ATTEMPTED because they neither confirm nor contradict the gold target.
- The gold target may contain more information than the question. In such cases, the predicted answer only needs to contain the information that is in the question.
    - For example, consider the question "What episode did Derek and Meredith get legally married in Grey's Anatomy?" with gold target "Season 7, Episode 20: White Wedding". Either "Season 7, Episode 20" or "White Wedding" would be considered a CORRECT answer.
- Do not punish predicted answers if they omit information that would be clearly inferred from the question.
    - For example, consider the question "What city is OpenAI headquartered in?" and the gold target "San Francisco, California". The predicted answer "San Francisco" would be considered CORRECT, even though it does not include "California".
    - Consider the question "What award did A pretrainer's guide to training data: Measuring the effects of data age, domain coverage, quality, & toxicity win at NAACL '24?", the gold target is "Outstanding Paper Award". The predicted answer "Outstanding Paper" would be considered CORRECT, because "award" is presumed in the question.
    - For the question "What is the height of Jason Wei in meters?", the gold target is "1.73 m". The predicted answer "1.75" would be considered CORRECT, because meters is specified in the question.
    - For the question "What is the name of Barack Obama's wife?", the gold target is "Michelle Obama". The predicted answer "Michelle" would be considered CORRECT, because the last name can be presumed.
- Do not punish for typos in people's name if it's clearly the same name. 
    - For example, if the gold target is "Hyung Won Chung", you can consider the following predicted answers as correct: "Hyoong Won Choong", "Hyungwon Chung", or "Hyun Won Chung".


Here is a new example. Simply reply with either CORRECT, INCORRECT, NOT ATTEMPTED. Don't apologize or correct yourself if there was a mistake; we are just trying to grade the answer.
```
Question: {question}
Gold target: {target}
Predicted answer: {predicted_answer}
```

Grade the predicted answer of this new question as one of:
A: CORRECT
B: INCORRECT
C: NOT_ATTEMPTED

Just return the letters "A", "B", or "C", with no text around it.
""".strip()
\end{lstlisting}

\section{Guessing strategy and F-score}
\label{sec:guessing_strategy}

While F-score is a good metric in some ways, the issue with it is that it incentivizes the model to always guess when it is at least 50\% sure that it can get the correct answer.
To understand why this is the case, consider the following expression for the F-score:
\[
F\textrm{-score} = \frac{2}{\frac{c + i}{c} + \frac{c + i + n}{c}} = \frac{2c}{2c + 2i + n}\ ,
\]
where:
\begin{itemize}
    \item $c$ is the number of correct answers,
    \item $i$ is the number of incorrect answers, and
    \item $n$ is the number of non-answered questions.
\end{itemize}

If you have a greater than $1/2$ chance of being correct, your expected score from guessing is better than the score from not guessing, regardless of the specific values for $c$, $i$, and $n$. This is because the following inequality always holds:

\[
\frac{2c + 1}{2c + 2i + n + 2} > \frac{2c}{2c + 2i + n + 1}\ .
\]

The left-hand side represents the expected F-score from guessing, assuming a 50/50 chance of correctness, while the right-hand side is the score from not answering the additional question. Since the denominators are adjusted similarly whether the guess is correct or incorrect, guessing with a probability $> 1/2$ yields a better score.

\section{Calibration Prompt}
\label{sec:calibration_prompt} 
\begin{lstlisting}[language={},frame=single,breaklines=true]
ANSWER_WITH_CONFIDENCE_PROMPT_TEMPLATE = """
Here is the question:
{question}
Please provide your best guess and a confidence score between 
0% to 100% in the following JSON format:
{
    "answer": "Your answer here",
    "confidence_score": number
}
""".strip()
\end{lstlisting}

\changeurlcolor{blue}
 
\end{document}